\documentclass[letterpaper]{article} 
\usepackage{aaai2026}  
\usepackage{times}  
\usepackage{helvet}  
\usepackage{courier}  
\usepackage[hyphens]{url}  
\usepackage{graphicx} 
\usepackage{natbib}  
\usepackage{caption} 
\usepackage{algorithm}
\usepackage{algorithmic}
\usepackage{amsmath}
\usepackage{xspace}
\usepackage{amsfonts}
\usepackage{booktabs}
\usepackage{multirow}
\usepackage{colortbl}
\usepackage{xcolor}
\usepackage{makecell}

\usepackage{newfloat}
\usepackage{listings}
\DeclareCaptionStyle{ruled}{labelfont=normalfont,labelsep=colon,strut=off} 
\floatstyle{ruled}
\newfloat{listing}{tb}{lst}{}
\floatname{listing}{Listing}
\title{Expand Your SCOPE: Semantic Cognition over Potential-Based Exploration\\for Embodied Visual Navigation}
\author{
    Ningnan Wang, Weihuang Chen\thanks{Corresponding author (chenwh@xjtu.edu.cn).}, Liming Chen, Haoxuan Ji,\\Zhongyu Guo, Xuchong Zhang, Hongbin Sun
}
\affiliations{
    State Key Laboratory of Human-Machine Hybrid Augmented Intelligence\\
    National Engineering Research Center for Visual Information and Applications\\
    Institute of Artificial Intelligence and Robotics, Xi'an Jiaotong University
}

\definecolor{lightgray}{gray}{0.9}

\begin{document}

\nocopyright
\maketitle

\begin{abstract}
Embodied visual navigation remains a challenging task, as agents must explore unknown environments with limited knowledge. Existing zero-shot studies have shown that incorporating memory mechanisms to support goal-directed behavior can improve long-horizon planning performance. However, they overlook visual frontier boundaries, which fundamentally dictate future trajectories and observations, and fall short of inferring the relationship between partial visual observations and navigation goals. In this paper, we propose Semantic Cognition Over Potential-based Exploration (SCOPE), a zero-shot framework that explicitly leverages frontier information to drive potential-based exploration, enabling more informed and goal-relevant decisions. SCOPE estimates exploration potential with a Vision-Language Model and organizes it into a spatio-temporal potential graph, capturing boundary dynamics to support long-horizon planning. In addition, SCOPE incorporates a self-reconsideration mechanism that revisits and refines prior decisions, enhancing reliability and reducing overconfident errors. Experimental results on two diverse embodied navigation tasks show that SCOPE outperforms state-of-the-art baselines by 4.6\% in accuracy. Further analysis demonstrates that its core components lead to improved calibration, stronger generalization, and higher decision quality.
\end{abstract} 

\begin{links}
    \link{Code}{https://github.com/mrwangyou/SCOPE}
    \link{Paper (AAAI 2026)}{https://ojs.aaai.org/index.php/AAAI/article/view/38929}
\end{links}

\section{Introduction}
Embodied visual navigation (EVN) is a fundamental task in embodied intelligence, requiring agents to autonomously plan paths toward designated goals in previously unseen environments. This capability holds significant promise for real-world applications such as smart home robotics, disaster response, and deep space exploration, and has attracted growing research interest in recent years~\cite{janny2025reasoning,ehsani2024spoc,wang2025instruction}.

Despite its utility, embodied visual navigation remains a significant challenge. 
Specifically, agents must operate in unfamiliar environments, continuously inferring associations between partial visual observations and navigation goals~\cite{li2025regnav}.
Furthermore, modern tasks increasingly involve multi-modal goal specifications, ranging from natural language commands to reference images or object categories, while demanding robust long-horizon planning~\cite{khanna2024goat}. 
These requirements place a premium on methods that combine strong semantic reasoning capabilities with reliable planning over extended sequences, substantially increasing decision-making complexity.

Strong exploration ability is essential to meeting these demands. Agents must efficiently acquire new information from current observations while effectively organizing and leveraging previously gathered knowledge. Most existing zero-shot navigation methods enhance exploration by incorporating memory mechanisms to support goal-directed behavior. For instance, ConceptGraph~\cite{gu2024conceptgraphs} constructs a graph-structured, object-centric representation of 3D scenes by fusing outputs from 2D foundation models, offering a compact memory for spatial reasoning. Similarly, VLFM~\cite{yokoyama2024vlfm} builds occupancy maps from depth data and employs a Vision-Language Model (VLM) to generate a language-grounded value map over frontiers. While such approaches have shown promise, they face two main limitations: 1) VLMs and LLMs, though proficient in multi-modal reasoning, are not specifically trained on dense 3D inputs, limiting their understanding of complex spatial relationships; and 2) memory systems focus solely on visited regions often fail to connect current observations to unvisited goal-relevant areas, leaving agents without clear exploration cues.

To address these limitations, recent methods have begun to incorporate information from unexplored regions. 
NaviFormer~\cite{xie2025naviformer} enhances object-goal navigation by integrating spatial layout encoding, agent pose trajectories, and a passable frontier map using spatio-temporal attention. 
3D-Mem~\cite{yang20253d}, the state-of-the-art method, proposes a snapshot-based 3D memory architecture that stores multi-view representations of explored areas alongside frontier snapshots of unexplored regions. 
However, these approaches still fall short in fully exploiting the \emph{potential} of visual frontiers. Far from being passive boundaries, frontiers act as critical bridges linking the agent's current state to potentially goal-relevant areas, shaping both future trajectories and the observations an agent can acquire.

This paper aims to propose a zero-shot method that is well-performing, robust, and capable of deep potential-based exploration. Specifically, we introduce Semantic Cognition Over Potential-based Exploration (SCOPE), a new approach that fully exploits the informational value of visual frontier boundaries to enhance long-horizon navigation. The key idea is to treat frontiers as semantic cognition cues that enhance scene understanding, memory construction, and decision reliability. SCOPE is composed of three key components. Firstly, a \emph{frontier-level potential estimator} evaluates the semantic relevance of unexplored regions with respect to the navigation goals from raw visual inputs. Secondly, a \emph{potential graph} constructs and dynamically updates a latent graph that propagates spatial-semantic utility as structured memory, guiding the VLM in decision-making. Finally, a \emph{self-reconsideration mechanism} revisits and refines initial decisions, preventing premature commitments, reducing hallucinations in VLM reasoning, and improving environmental understanding and planning accuracy. These components work in synergy to foster more deliberate and informed navigation, leading to stronger performance across complex tasks.

Our main contributions are as follows:
\begin{enumerate}
    \item We propose a zero-shot embodied navigation framework that explicitly prioritizes frontier information as a primary exploration cue, introducing a new paradigm for information-guided exploration.
    \item We design a potential graph that efficiently integrates spatio-temporal information with frontier potential estimation, enabling structured environment representation and informed global planning.
    \item We incorporate a self-reconsideration mechanism, encouraging iterative refinement and contextual alignment, leading to more reliable and deliberate actions.
    \item We demonstrate statistically significant improvements over state-of-the-art baselines, achieving \textbf{4.6\%} accuracy gains on two challenging embodied navigation benchmarks (GOAT-Bench and A-EQA). 
\end{enumerate}
Further analysis reveals that our method delivers statistically significant gains, demonstrates better-calibrated decision confidence, and benefits from each proposed component. We release the complete implementation of SCOPE for transparency and reproducibility.

\begin{figure*}[t]
    \centering
    \includegraphics[width=0.89\textwidth]{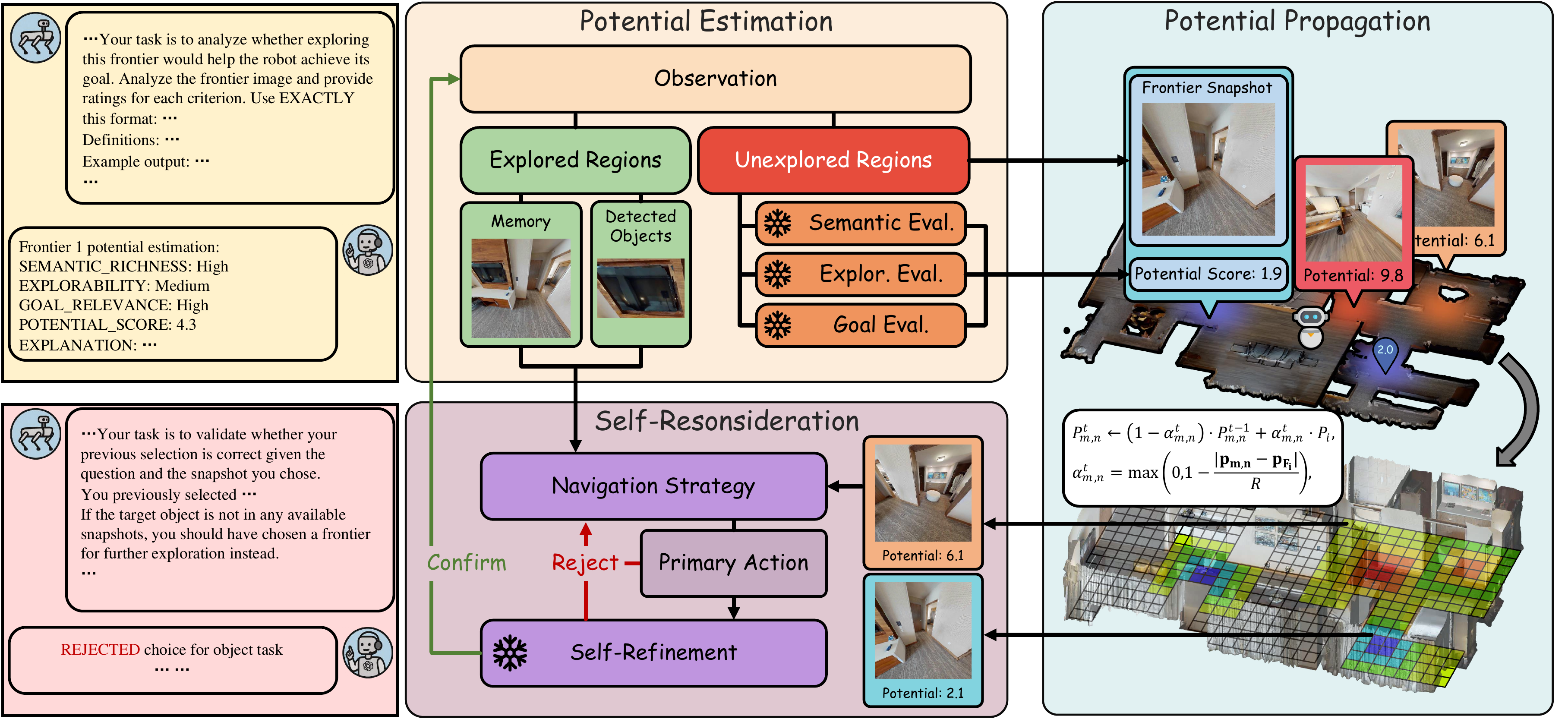}
    \caption{Overview of SCOPE. The agent predicts frontier utility via a VLM-based estimator and encodes it into a structured potential graph for spatiotemporal reasoning. Action decisions are guided by this graph and further reconsideration through a self-refinement module to avoid impulsive errors.}
    \label{fig:main}
\end{figure*}

\section{Related Work}

\subsubsection{EVN Tasks}
EVN tasks involve a wide range of task formulations. ObjectNav directs agents to locate an instance of a specified semantic category~\cite{zhang2024imagine,khanna2024habitat,yin2024sg,zhang2025hoz++}. ImageNav requires finding an object that visually matches a given reference image~\cite{barsellotti2024personalized,li2025regnav,lei2024instance,lei2025gaussnav}. Vision-Language Navigation (VLN) requires agents to ground semantic descriptions into spatial targets~\cite{perincherry2025visual,wang2024lookahead,wang2024vision,an2024etpnav}. Beyond these forms, Embodied Question Answering (EQA) requires agents to explore environments to collect information needed to answer natural-language questions~\cite{RenCDIMS24,wang2025embodied,dang2025ecbench}. Emerging multi-stage~\cite{song2025towards} and multi-step~\cite{yu2024trajectory} navigation settings further amplify complexity by demanding extended reasoning, persistent memory, and adaptive long-horizon planning, which are core capabilities forming the focus of this work.

\subsubsection{Zero-shot EVN Methods}
Zero-shot EVN methods aim to enable agents to operate in unknown environments without task-specific fine-tuning, instead leveraging generalizable knowledge learned from large-scale pretraining. These approaches have recently gained significant attention and generally progress along two main directions. The first focuses on improving scene representation and memory organization to better support spatial reasoning and long-term planning~\cite{liu2024volumetric,yin2025unigoal,zhao2024over,li2024memonav,lin2025bip3d,xie2025naviformer}. The second emphasizes enhancing agent reasoning through advanced decision-making paradigms, such as chain-of-thought prompting~\cite{lin2025navcot,kong2024controllable}, diffusion-based decision policies~\cite{ren2025prior,zhang2024versatile}, or world models~\cite{bar2025navigation,pan2025planning} for simulating and planning in dynamic environments. While these methods often excel at modeling visited regions, they lack structured and adaptive mechanisms for prioritizing which unexplored areas to visit next. In contrast, our approach treats frontiers as a primary semantic and structural cue for exploration, enabling more targeted, informed, and goal-aligned navigation.

\subsubsection{Frontier-related EVN Methods}
Several recent works have explicitly incorporated frontier information into EVN policies.  
VLFM~\cite{yokoyama2024vlfm} derives semantically valuable cues from occupancy maps to guide exploration toward unseen object targets. 
\citeauthor{cui2024frontier}~(\citeyear{cui2024frontier}) represent visible but unexplored frontiers with ``ghost nodes'' to improve exploration awareness in navigation agents. However, both methods rely mainly on simple geometric frontier features, limiting adaptability to diverse navigation contexts. 3D‑Mem~\cite{yang20253d} stores snapshots of unexplored regions, enabling VLM-based reasoning over frontier areas. 
While this extends frontier usage beyond geometric estimation, the approach largely passes raw snapshots to VLMs and still depends on heuristic frontier selection, leaving much of the potential of frontiers untapped. In contrast, our method introduces a potential-based frontier exploration paradigm that explicitly estimates the semantic utility of each frontier, integrates it into a structured spatio-temporal potential graph, and refines decisions via self-reconsideration.

\section{Methodology}

\subsection{Preliminary}
EVN tasks require an agent to traverse an unknown environment to reach goals specified by point coordinates, object references, or natural language instructions. 
Formally, at each time step $t$, the agent selects an action $a^t$ using a navigation policy $\pi_\theta$, conditioned on the current goal $q$, the agent's state $s^t$, and its accumulated knowledge $\mathcal{K}^t$:
\begin{equation}
a^t = \pi_\theta(q, s^t, \mathcal{K}^t),
\end{equation}
where $\mathcal{K}^t = \{I^1, I^2, \dots, I^t\}$ denotes the sequence of historical observations up to time $t$. Upon executing $a^t$, the agent transitions to a new state $s^{t+1}$ and receives new sensory input. If the goal is deemed satisfied, the episode terminates, the task is considered finished, and a goal checker would examine whether the goal is achieved. Otherwise, the agent continues exploring to collect additional information.

At each step, the agent captures a panoramic set of egocentric observations:
\begin{equation}
I^t = \{I_1^t, I_2^t, \dots\},
\end{equation}
where each $I_i^t$ is an RGB-D image from a discrete directional view. From these observations, the agent constructs an \emph{explored memory} $\mathcal{E}^t$, which encodes both the raw visual snapshot and detected semantic entities:
\begin{equation}
\mathcal{E}^t = \{I^t, O^t\},
\end{equation}
with $O^t$ representing object-level annotations or region semantics extracted from $I^t$.

In parallel, the agent identifies \emph{frontier regions}---boundaries between explored and unobserved space. These frontiers are represented as spatial clusters $\mathcal{F}^t = \{\mathcal{F}_1^t, \mathcal{F}_2^t, \dots\}$ at the periphery of explored areas, each associated with a representative visual snapshot.

We define the structured environmental knowledge $\tilde{\mathcal{K}}^t$ as the union of all previously explored content and current frontier candidates:
\begin{equation}
\tilde{\mathcal{K}}^t = \underbrace{\{\mathcal{E}^1, \mathcal{E}^2, \dots, \mathcal{E}^t\}}_{\text{Explored regions}} \cup \underbrace{\mathcal{F}^t}_{\text{Frontier regions}}.
\end{equation}
Therefore, the environment is partitioned into two complementary components: explored regions, which contain visited areas with available visual and semantic information, and frontier regions, which represent unobserved boundary areas serving as gateways to new information. Our framework explicitly models and exploits frontier regions as informative cues for decision-making, enabling goal-directed and efficient exploration in unknown areas.


\subsection{Framework}

The overall framework of SCOPE is illustrated in Fig.~\ref{fig:main}. The agent first captures observations encompassing both explored and frontier regions. To fully leverage information from frontier observations, we introduce a frontier-level potential estimator built upon open-domain VLMs, which integrates visual features from frontier snapshots with task-specific contextual cues to predict the semantic utility of each unexplored region. These predicted utilities are then incorporated into a potential graph that encodes, propagates, and updates utility estimates across spatial and temporal dimensions, enabling the agent to prioritize high-utility frontiers, revisit semantically valuable areas, and avoid low-utility regions to improve efficiency. The enriched frontier snapshots, now carrying propagated utility estimates, are fed back into the agent's decision-making loop to guide more informed action selection. To further ensure planning stability and mitigate erroneous or impulsive choices, we incorporate a self-reconsideration mechanism that performs internal consistency checks on candidate actions before execution, allowing the agent to correct flawed reasoning and refine its strategy based on accumulated knowledge.

\subsection{Frontier-level Potential Estimation}

To assess the utility of each unexplored frontier, we propose a frontier-level potential estimator grounded in pretrained VLMs. This estimator jointly leverages visual features from frontier snapshots and goal-specific prompts to infer semantic utility, effectively transforming the VLM into a generalizable semantic oracle across scenes and tasks.


Specifically, the potential of an unexplored region can be assessed from three complementary perspectives. Semantic Richness: Regions with high semantic density would efficiently earn more information and are more likely to include goal-relevant content. Explorability: A valuable frontier should not only be informative itself but also serve as a gateway to other unseen areas. Goal Relevance: Semantically related objects often co-occur in the same place. A frontier showing partial evidence of such a semantic cluster increases the likelihood that the goal target lies nearby.

Formally, for each frontier snapshot $\mathcal{F}_i^t \in \mathcal{F}^t$, we define:
\begin{equation}
\mathbf{p}_i^t=[p_{i,sem}^t,p_{i,explore}^t,p_{i, goal}^t]=f_{\text{VLM}}(\mathcal{F}_i^t,q),
\end{equation}
where $f_{\text{VLM}}$ is a prompting-based interface to a VLM, and $q$ is the task goal. The model also provides an aggregated scalar score:
\begin{equation}
P_i^t=\text{Aggregate}(\mathbf{p}_i^t),
\end{equation}
indicating the overall utility of each frontier snapshot. 

This component transforms the VLM into a goal-aware semantic oracle that generalizes across tasks and scenes without additional training. It grounds frontier evaluation in rich pretrained priors and enables interpretable, context-aware decision-making.

\subsection{Structured Memory over Potential Graph}

To enable long-horizon planning and mitigate aimless exploration, we introduce the potential graph, a spatial memory structure that encodes the estimated utility of different regions across the navigable environment. This graph captures past frontier observations, propagates their potential across space, and enables strategic revisitation or avoidance.

We discretize the environment into a 2D grid $\mathcal{G} = \{v_{m,n}\}_{m,n}$, where each cell $v_{m,n}$ corresponds to a spatial location and maintains the properties including a potential score $P_{m,n}$, a visit count $n_{m,n}$, and semantic attributes $p_{m,n}^{\text{sem}},\ p_{m,n}^{\text{explore}},\ p_{m,n}^{\text{goal}}$.

When a frontier $\mathcal{F}_i$ is observed at position $\mathbf{p}_{m,n}$ with estimated potentials $s_i = \left( p_i^{\text{sem}},\ p_i^{\text{explore}},\ p_i^{\text{goal}},\ P_i \right)$, its information is propagated to nearby graph nodes within a fixed radius $R$. For each affected cell $v_{m,n}$ with spatial coordinate $\mathbf{p}_{m,n}$, we perform the following weighted update:
\begin{equation}
P_{m,n}^{t} \leftarrow (1 - \alpha_{m,n}^{t}) \cdot P_{m,n}^{t-1} + \alpha_{m,n}^{t} \cdot P_i,
\end{equation}
\begin{equation}
\alpha_{m,n}^{t} = \max \left(0,1 - \frac{\| \mathbf{p}_{m,n} - \mathbf{p}_{\mathcal{F}_i} \|}{R} \right),
\end{equation}
where $\alpha_{m,n}^t \in [0,1]$ determines the influence weight based on the spatial distance between the frontier and the cell. 
A similar weighted update is applied to semantic components $p^{\text{sem}},\ p^{\text{explore}},\ p^{\text{goal}}$.

To encourage wide coverage and avoid local cycles, we compute a frontier exploration value for each node by combining its spatial potential and semantic relevance, while penalizing frequent revisits. Specifically, a weighted sum of the node's potential score, along with its estimated semantic richness, explorability, and goal relevance assessments, is computed using a weight vector $\mathbf{w}=[\omega_{\text{pot}}, \omega_{\text{sem}},\omega_{\text{explore}},\omega_{\text{goal}}]$. This semantic score is then scaled by an inverse visitation factor:
\begin{equation}
\begin{split}
E_{m,n} = \big( &\omega_{\text{pot}} \cdot P_{m,n} 
+ \omega_{\text{sem}} \cdot p_{m,n}^{\text{sem}} \\
&+ \omega_{\text{explore}} \cdot p_{m,n}^{\text{explore}} 
+ \omega_{\text{goal}} \cdot p_{m,n}^{\text{goal}} \big) \cdot \frac{1}{1 + \gamma \cdot \mathbf{n}_{m,n}},
\end{split}
\end{equation}
where $\mathbf{n}_{m,n}$ is the visit count of node $v_{m,n}$ and $\gamma > 0$ is a decay coefficient controlling the strength of revisitation penalty. This scoring encourages selection of novel yet promising areas. During navigation, the agent is prompted with the estimated potential score and consults the active frontier scores to plan its next move, facilitating strategic long-term exploration and reducing inefficient behaviors.

\subsection{Self-reconsideration for Robust Action Validation}

To avoid premature and inaccurate decisions, we incorporate a self-refinement technique that validates candidate actions before execution. This technique assesses whether a selected memory snapshot truly contains the goal-relevant content.

\subsubsection{Refinement Trigger}

Given a goal query $q$ and structured memory $\tilde{\mathcal{K}}^t$, the agent first produces a primary action:
\begin{equation}
a_t^{(0)} = \pi_\theta(q, \tilde{\mathcal{K}}^t),
\end{equation}
which corresponds to either selecting a memory snapshot and object pair or proposing a frontier snapshot to explore. If this action selects a memory snapshot-object pair, refinement is triggered:

\begin{equation}
\delta(a_t^{(0)}) = 
\begin{cases}
1 & \text{if } a_t^{(0)} = \texttt{memory}(i, o), \\
0 & \text{if } a_t^{(0)} = \texttt{frontier}(j).
\end{cases}
\end{equation}

\subsubsection{Validation Process}

If refinement is triggered, a secondary decision is made by invoking a validation model $g_\phi$, using a VLM, to assess whether the selected snapshot actually satisfies the task goal:
\begin{equation}
r_t^{(0)} = g_\phi(q, \mathcal{E}, o),
\end{equation}
where $r_t \in \{\texttt{CONFIRM}, \texttt{REJECT}\}$. The input to $g_\phi$ includes the task goal $q$, the selected snapshot $\mathcal{E} = \langle I_{\mathcal{E}}, O_{\mathcal{E}} \rangle$, and the object $o \in O$ that was chosen as the goal candidate.

If $r_t^{(0)} = \texttt{CONFIRM}$, the original action $a_t^{(0)}$ is accepted as the final action $a_t = a_t^{(0)}$.
If $r_t^{(0)} = \texttt{REJECT}$, the agent discards the selection and re-consults the main policy:
\begin{equation}
a_t^{(1)} = \pi_\theta(q, \tilde{\mathcal{K}}^t, a_t^{(0)}),
\end{equation}
possibly repeating this process with updated guidance until either a valid action is confirmed or a retry limit is reached.

\subsubsection{Final Action Selection}

Overall, the agent's action at time $t$ is determined as:

\begin{equation}
    a_t=\text{SR}(a_t^{(0)}),
\end{equation}
\begin{equation}
\text{SR}(a_t^{(i)}) =
\begin{cases}
a_t^{(i)} & \text{if } \delta(a_t^{(i)}) = 0 \text{ or } r_t^{(i)} = \texttt{CONFIRM}, \\
a_t^{(i+1)} & \text{if } r_t^{(i)} = \texttt{REJECT}.
\end{cases}
\end{equation}

This forms a corrective feedback loop that allows the agent to reassess decisions and avoid overconfident mistakes, improving robustness in ambiguous scenarios.

\section{Experiments}

\subsection{Experiment Settings}

\subsubsection{Benchmarks}
To enable a broad evaluation across key embodied reasoning challenges, we leverage two complementary benchmarks covering distinct tasks: GOAT-Bench~\cite{khanna2024goat}, which targets goal-conditioned navigation grounded in diverse language instructions, and A-EQA~\cite{majumdar2024openeqa}, which emphasizes embodied question answering that demands multi-step spatial and semantic reasoning. These benchmarks reflect distinct task formulations---spatial navigation and semantic reasoning---making them ideal to test the generality and robustness of our method.

\subsubsection{Metrics}

We adopt four established metrics to comprehensively evaluate our agent's performance across navigation and embodied question answering tasks. For GOAT-Bench, we follow prior work in using Success Rate (SR) and Success weighted by Path Length (SPL). SR measures the percentage of subtasks in which the agent stops within 1 meter of the correct goal object, while SPL accounts for both success and trajectory efficiency by penalizing unnecessarily long paths. For A-EQA, we evaluate answer quality using Correctness, based on the LLM-Match score---a large language model-based metric that aligns closely with human judgment. To assess information-seeking behavior, we also report Efficiency, defined as the ratio between the shortest possible path to the relevant information and the actual path the agent traverses before answering. Together, these four metrics provide a holistic view of agent performance in terms of task success, path optimality, answer quality, and exploration efficiency.

\subsubsection{Baselines}

To ensure a fair and comprehensive comparison, we benchmark our method against 3D-Mem~\cite{yang20253d}, the state-of-the-art framework which introduces a spatial memory module to support long-term, goal-directed reasoning. By persistently storing visual and semantic information throughout exploration, 3D-Mem enables effective retrieval and reasoning over past observations, making it a strong representative baseline for open-ended embodied navigation. 
In addition to 3D-Mem, we evaluate against several recent and competitive methods. 
For example, Explore-EQA~\cite{ren2024explore} integrates semantic-guided frontier exploration with a calibrated stopping mechanism based on conformal prediction.
TANGO~\cite{ziliotto2025tango} is a training-free framework that assembles pre-trained vision and language models into agents capable of solving open-world tasks, demonstrating strong generalization without task-specific training. MTU3D~\cite{zhu2025move} proposes a unified framework that bridges visual grounding and exploration by jointly optimizing scene understanding and navigation policies within a memory-augmented transformer architecture. Together, these baselines represent diverse and advanced strategies for perception, memory, and reasoning, providing strong reference points for evaluating the effectiveness of our proposed approach.

\subsubsection{Experimental Details} 
We utilize the GPT-4o API to handle VLM requests, as it achieves the best performance when integrated into our baseline agents, outperforming Gemini-2.0-Flash, Pixtral-Large, Qwen-Omni-Turbo, and LLaVA-1.5-13B. 
All experiments are conducted on a server equipped with 8×NVIDIA A800 80GB GPUs.

To ensure full reproducibility and transparency, all code and exact hyperparameters used in our experiments are publicly available.
Additional implementation details can be found in Appendix.

\subsection{Results}

\begin{table}[t]
    \centering
    \begin{tabular}{lcccc}
         \toprule[1.5pt]
          \multicolumn{5}{c}{\textbf{GOAT-Bench}} \\
          \textbf{Method} & FR & FI & \textbf{SR} & \textbf{SPL} \\
          \midrule[0.75pt]
          {SenseAct-NN Skill Chain} & -- & -- & 29.5 & 11.3\\
          {SenseAct-NN Monolithic} & -- & --& 12.3 & 6.8\\
          {CoW}~(\citeyear{gadre2023cows}) &-- &-- & 16.1 & 10.4 \\
          {DyNaVLM}~(\citeyear{ji2025dynavlm}) & --& --& 25.5 & 10.2 \\
          {VLMnav}~(\citeyear{goetting2024endtoend}) & --& --& 16.3 & 6.6 \\
          {Modular GOAT}~(\citeyear{khanna2024goat}) & --& --& 24.9 & 17.2 \\
          
          \midrule[0.75pt]
          {Explore-EQA}~(ICRA'\citeyear{ren2024explore}) & \checkmark & --& 55.0 & 37.9 \\
          {CG + Frontier}~(ICRA'\citeyear{gu2024conceptgraphs}) & \checkmark &  --& 61.5 & 41.3 \\
          {TANGO}~(CVPR'\citeyear{ziliotto2025tango}) &-- &-- & 32.1 & 16.5 \\
          {MTU3D}~(ICCV'\citeyear{zhu2025move}) & \checkmark & -- & 47.2 & 27.7\\
          {3D-Mem}~(CVPR'\citeyear{yang20253d}) & \checkmark & --&{69.1} &{48.9}\\ \midrule[0.75pt]
          {SCOPE}~(Ours) & \checkmark & \checkmark & \textbf{73.7} & \textbf{53.5} \\
          \bottomrule[1.5pt]
    \end{tabular}
    \caption{Performance comparison on \textbf{GOAT-Bench}. \textbf{Bold} indicates the best performance.}
    \label{tab:metricsgoat}
\end{table}

\begin{figure}[b]
    \centering
    \includegraphics[width=0.9\columnwidth]{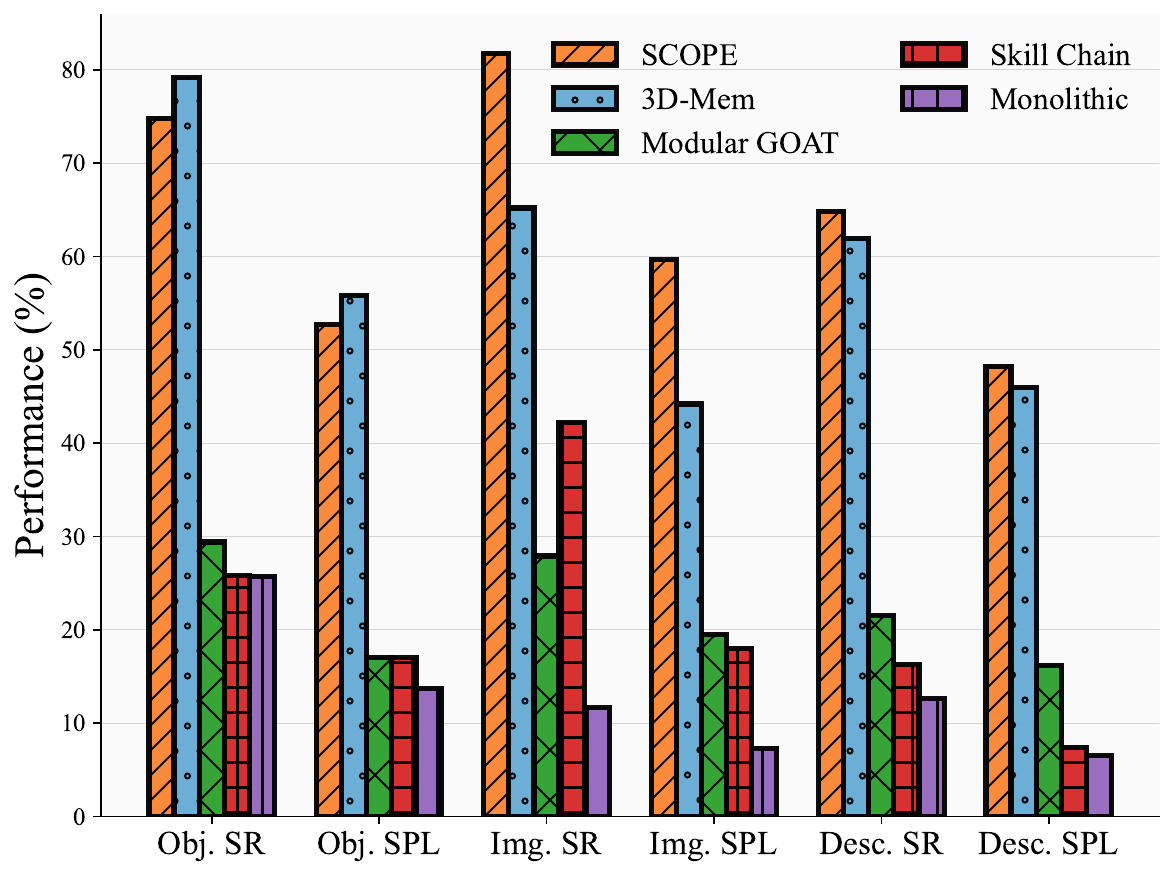} 
    \caption{Performance across modalities on GOAT-Bench.}
    \label{fig:category}
\end{figure}

We evaluate SCOPE on vision-language navigation benchmark GOAT-Bench, and the result is presented in Table~\ref{tab:metricsgoat} and Fig.~\ref{fig:category}. 
FR denotes methods that incorporate frontier representations, while FI refers to methods that fully leverage frontier information for decision-making.
SCOPE achieves the highest SR on GOAT-Bench, improving over the previous most competitive method, 3D-Mem, by +4.6\% (73.7 vs. 69.1). Despite being more accurate, SCOPE also improves efficiency. It achieves a higher SPL by +4.6\% (53.5 vs. 48.9).

\begin{table}[t]
    \centering
    \begin{tabular}{lcccc}
         \toprule[1.5pt]
          \multicolumn{5}{c}{\textbf{A-EQA}} \\
          \textbf{Method} & FR & FI &  \textbf{Corr.} & \textbf{Eff.} \\
          \midrule[0.75pt]
          {LLaVA-1.5 Frame Captions} & --& --& 38.1 & 7.0 \\
          {CG Scene-Graph Captions} & --& --& 34.4 & 6.5  \\
          {SVM Scene-Graph Captions} & --& --& 34.2 & 6.4 \\
          {Multi-Frame} & --& --& 41.8 & 7.5 \\
          \midrule[0.75pt]
          {Explore-EQA}~(ICRA'\citeyear{ren2024explore}) & \checkmark & --& 46.9 & 23.4 \\
          {CG + Frontier}~(ICRA'\citeyear{gu2024conceptgraphs})  & \checkmark & --& 47.2 & 33.3 \\
          {TANGO}~(CVPR'\citeyear{ziliotto2025tango}) & --& --& 37.2 & / \\
          {MTU3D}~(ICCV'\citeyear{zhu2025move}) & \checkmark & --& 51.1 & \textbf{42.6}\\
          {3D-Mem}~(CVPR'\citeyear{yang20253d}) & \checkmark & --& {52.6} &{42.0} \\
          \midrule[0.75pt]
          {SCOPE}~(Ours)& \checkmark & \checkmark & \textbf{59.1} & 41.0 \\
          \bottomrule[1.5pt]
    \end{tabular}
    \caption{Performance comparison on \textbf{A-EQA}. \textbf{Bold} indicates the best performance. ``/'' denotes unreported results.}
    \label{tab:metricseqa}
\end{table}

We also evaluate SCOPE on A-EQA, an embodied question answering benchmark, and the result is shown in Table~\ref{tab:metricseqa}. On A-EQA, SCOPE boosts answer accuracy by +6.5\% over 3D-Mem (59.1 vs. 52.6), demonstrating better semantic comprehension and task alignment. Compared to MTU3D (the most efficient prior method on A-EQA with efficiency 42.6), SCOPE slightly trades off efficiency (41.0 vs. 42.6) for a notable gain in answer accuracy (+8.0\%, 59.1 vs. 51.1), suggesting better generalizability and reliability. These results underscore the utility of our utilization of frontier information in enhancing long-term decision quality.

\section{Analysis}

To further evaluate the effectiveness of SCOPE and the contributions of our principle of fully utilizing frontier information, we conduct a series of analyses beyond standard evaluation. We evaluate the significance and calibration of SCOPE's performance, and further isolate the contributions of each core component through targeted studies on potential estimation, potential graph propagation, and self-refinement.

\subsection{Statistical Significance Analysis}

To rigorously evaluate the performance improvements of SCOPE over the strongest baseline, we conduct a statistical significance analysis against the state-of-the-art method 3D-Mem. Both methods are independently run five times under identical random seed settings, and we report their mean success rates along with standard deviations to capture variability across runs.

\begin{figure}[t]
    \centering
    \includegraphics[width=0.98\columnwidth]{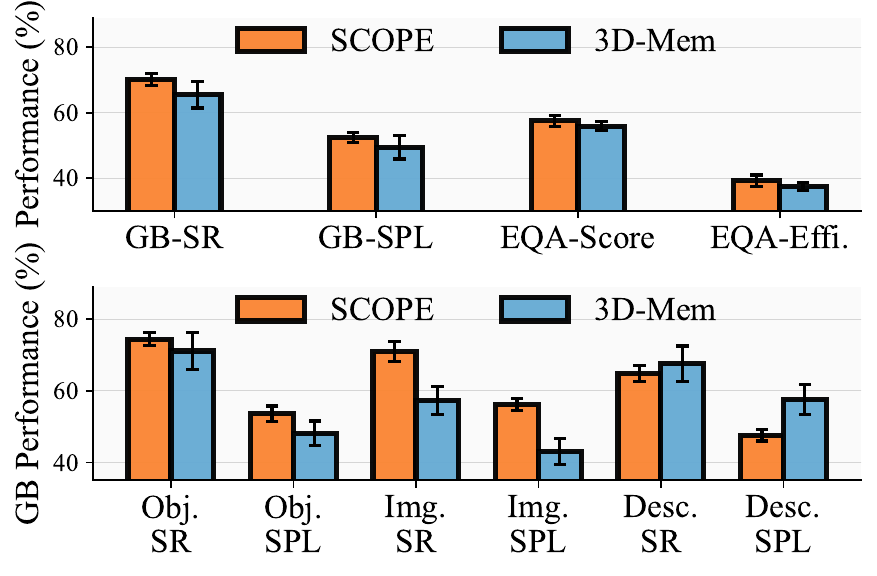} 
    \caption{Performance comparison between SCOPE and 3D-Mem. \textbf{Top}: Results on the GOAT-Bench and A-EQA benchmarks, covering goal-based navigation (GB) and embodied question answering (EQA) tasks. \textbf{Bottom}: Detailed breakdown of GOAT-Bench SR and SPL across object-, image-, and description-goal settings. SCOPE achieves higher average performance and lower variance than 3D-Mem.}
    \label{fig:significance}
\end{figure}

The results are shown in Fig.~\ref{fig:significance}, 
we apply an unpaired two-tailed t-test to assess the statistical significance of these improvements. The resulting p-value on GOAT-Bench is 0.046, indicating that the improvement is statistically significant at the 5\% level. On A-EQA, the p-value is 0.1365, suggesting a positive trend in favor of SCOPE.
These results indicate that SCOPE not only achieves higher average performance but also exhibits lower variance across runs. This highlights the robustness and consistency of its improvements over 3D-Mem. Together with the calibration analysis, the findings support SCOPE as a more stable and effective agent for embodied navigation tasks.

\begin{figure}[h]
    \centering
    \includegraphics[width=0.9\columnwidth]{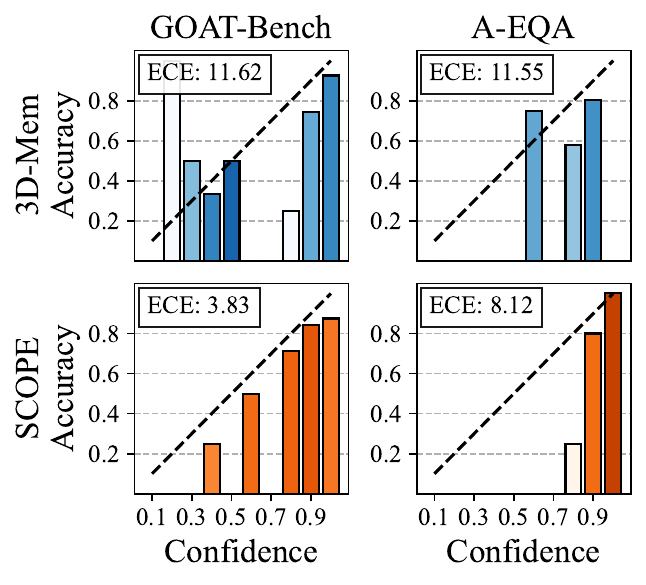} 
    \caption{Calibration of 3D-Mem and SCOPE. ``ECE'' represents the estimated calibration error ($\times 100$), with lower values indicating better calibration. The dashed line denotes perfect calibration, and the bar colors become darker as they approach ideal calibration.}
    \label{fig:calibration}
\end{figure}

\subsection{Model Calibration}

Reliable navigation in embodied agents requires not only accurate decisions but also well-calibrated confidence scores that reflect the agent's uncertainty---crucial for downstream decision-making and real-world deployment. Calibration measures how closely a model's predicted confidence aligns with the true likelihood of correctness~\cite{guo2017calibration}. A well-calibrated agent enables more trustworthy predictions and better risk assessment in practical deployments, which is especially important for an embodied agent.

To evaluate the reliability of SCOPE, we perform a calibration analysis on GOAT-Bench and A-EQA. We compute Expected Calibration Error (ECE) to quantify the discrepancy between predicted confidence and actual accuracy. 
As shown in Fig.~\ref{fig:calibration}, SCOPE achieves superior calibration performance. On GOAT-Bench, it reduces ECE from 11.62 (of the most competitive baseline 3D-Mem) to 3.83, and on A-EQA, from 11.55 to 8.12. These results indicate that SCOPE not only improves task performance but also produces more trustworthy estimates, equipping the agent with more reliable self-assessment for acting under uncertainty.

\subsection{Utility Analysis}

To further examine the utility of our utilization of the frontier, we design analysis experiments to each of our proposed component and evaluate its effectiveness.

\subsubsection{Utility of Potential Estimator}
As for the proposed potential estimator, we design an analysis experiment that alters the form of the frontier input provided to the agent while keeping the estimated potential scores fixed. 
Specifically, we discard the original frontier images and instead provide the agent with a brief textual description of each frontier region, while still supplying the estimated potential score.
This setup isolated the role of the potential scores as an independent guiding signal: if the agent can still navigate competently without visual frontier input, it indicates that the estimator encodes high-level, semantically meaningful, and goal-aware information beyond what is directly observable from raw pixels.

\begin{figure}[t]
    \centering
    \includegraphics[width=0.9\columnwidth]{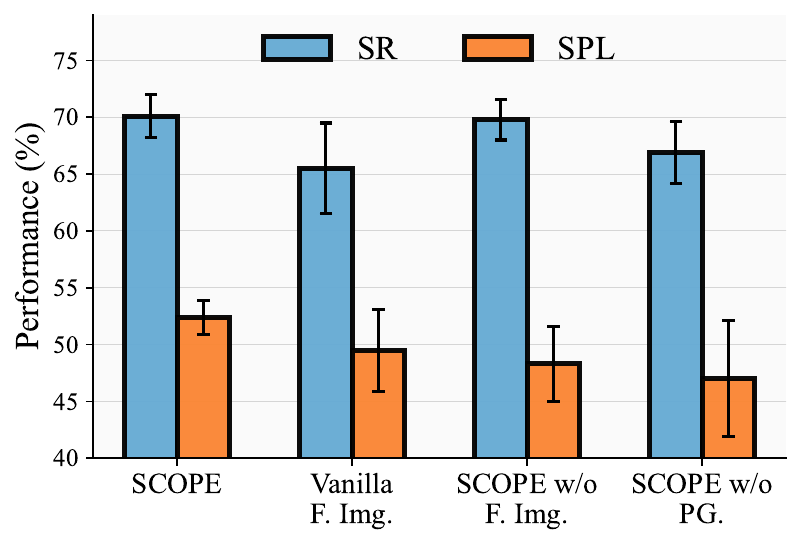}
    \caption{Ablation study evaluating the contribution of SCOPE components. \textbf{SCOPE w/o F. Img.} removes the frontier image input to the agent while retaining it for the potential estimator. \textbf{SCOPE w/o PG.} disables the potential graph module, exposing the agent only to raw estimated potential scores without spatial propagation.}
    \label{fig:ablation}
\end{figure}

The results are shown in Fig.~\ref{fig:ablation}. Remarkably, even without frontier images, the agent maintains strong performance---achieving results on par with the full SCOPE model and significantly outperforming vanilla utilization of frontier images. This demonstrates that our potential estimation captures robust and task-relevant frontier representations that generalize beyond the visual modality. The potential scores alone equip the agent with a sufficiently rich understanding of the environment, enabling it to make informed decisions regardless of how the frontier is represented. These findings underscore the versatility and strength of our estimator in supporting high-level reasoning and navigation.

\subsubsection{Utility of Potential Spread}

To assess the impact of the potential graph, we perform an ablation study where the agent only has access to the latest estimated potential scores of candidate frontiers, without the structured spatial-semantic propagation maintained by the potential graph. Unlike the full SCOPE model that accumulates and diffuses utility information over time and space, this variant makes decisions based solely on instantaneous frontier potentials, ignoring historical and contextual information.

As shown in Fig.~\ref{fig:ablation}, removing the potential graph leads to a clear performance drop: SR decreases from about 70.1\% to 66.9\%, and SPL falls from 52.4\% to 47.0\%. This result highlights the importance of structured spatial memory for long-term planning, allowing the agent to recall and leverage prior knowledge to revisit promising areas and avoid inefficient behaviors. The potential graph thus plays a crucial role in improving navigation efficiency and robustness beyond what raw frontier estimation alone can achieve.

\subsubsection{Utility of Re-consideration}

To evaluate the utility of the proposed self-refinement technique, we conduct both quantitative and qualitative analyses. We begin by aggregating the outcomes of refinement decisions across all episodes. When the refinement module confirms an answer, it is correct in 80.6\% of cases. Conversely, when the module consider an answer needing reconsideration, yet the answer is ultimately executed due to reaching the retry limit, it turns out to be incorrect in 34.3\% of cases. These results demonstrate the effectiveness of self-refinement in both reinforcing correct judgments and mitigating overconfident errors.


\section{Conclusion}

We presented SCOPE, a new framework for embodied visual navigation tasks that explicitly leverages visual boundaries as a central cue for informed exploration and long-term planning. By estimating frontier potential and organizing it within a structured potential graph, SCOPE enables agents to reason more effectively about where to explore next. The addition of a self-reconsideration mechanism further enhances decision reliability by revisiting and refining prior actions. Empirical results confirm that SCOPE outperforms state-of-the-art baselines and improves calibration, generalization, and decision quality. These findings highlight the importance of structured, frontier-aware reasoning in advancing goal-driven embodied navigation.

\section*{Acknowledgments}
This research was supported by the National Natural Science Foundation of China (No. 92464302, No.62503381, No. U24A20291).

\appendix

\bibliography{aaai2026}

\section{Reproducibility}

Our implementation is built on the 3D-Mem repository. We provide the SCOPE code along with all (hyper)parameters to ensure reproducibility. We also present some of the (hyper)parameters in Table~\ref{tab:hyperparameter}.

\begin{table}[h]
    \centering
    \begin{tabular}{lc}
         \toprule[1.5pt]
           (Hyper)parameters & Value \\
          \midrule[0.75pt]
          \textit{Sensor-related hyperparameters} \\
          camera\_height   & 1.5 \\
          camera\_tilt\_degree & -30 \\
          img\_width & 1280 \\
          img\_height & 1280 \\
          HFOV & 120 \\
          success\_distance & 1.0 (m) \\
          \midrule[0.75pt]
          \textit{VLM-related hyperparameters} \\
          GPT model & gpt-4o-2024-11-20 \\
          prompt\_img\_width & 360 \\
          prompt\_img\_height & 360 \\
          max\_tokens & 4096 \\
          top\_p & 0.95 \\
          frequency\_penalty & 0 \\
          presence\_penalty & 0 \\

          \bottomrule[1.5pt]
    \end{tabular}
    \caption{(Hyper)parameters of SCOPE.}
    \label{tab:hyperparameter}
\end{table}

\section{Full-set Evaluation}

To validate the robustness and effectiveness of SCOPE, we first evaluate its performance across the complete GOAT-Bench, which encompasses a broad spectrum of indoor navigation challenges with diverse semantic and spatial complexities. The comprehensive results, summarized in Table~\ref{tab:wholeset}, demonstrate that SCOPE consistently outperforms prior baselines across multiple goal types and task difficulties.

\begin{table}[h]
    \centering
    \begin{tabular}{lcccc}
         \toprule[1.5pt]
         \multirow{2}{*}{\textbf{Full Run}} & \multicolumn{2}{c}{\textbf{Success Rate}} & \multicolumn{2}{c}{\textbf{SPL}} \\
          & \textbf{Whole set} & \textbf{Subset} & \textbf{Whole set} & \textbf{Subset} \\
          \midrule[0.75pt]
          3D-Mem & 62.9 & 69.1 & 44.7 & 48.9 \\ 
          SCOPE & \textbf{66.8} & \textbf{73.7} & \textbf{46.5} & \textbf{53.5} \\
          \midrule[0.75pt]
          \textbf{5x GB} & \multicolumn{2}{c}{\textbf{Success Rate}} & \multicolumn{2}{c}{\textbf{SPL}}\\
          \midrule[0.75pt]
          3D-Mem & \multicolumn{2}{c}{65.5 $\pm$ 4.0} & \multicolumn{2}{c}{49.5 $\pm$ 3.6}\\
          SCOPE & \multicolumn{2}{c}{\textbf{70.1 $\pm$ 1.9}} & \multicolumn{2}{c}{\textbf{52.4 $\pm$ 1.5}}\\
          \midrule[0.75pt]
          \textbf{5x EQA} & \multicolumn{2}{c}{\textbf{Score}} & \multicolumn{2}{c}{\textbf{Efficiency}}\\
          \midrule[0.75pt]
          3D-Mem & \multicolumn{2}{c}{55.9 $\pm$ 1.4} & \multicolumn{2}{c}{37.5 $\pm$ 1.2}\\
          SCOPE & \multicolumn{2}{c}{\textbf{57.5 $\pm$ 1.6}} & \multicolumn{2}{c}{\textbf{39.2 $\pm$ 1.8}}\\
          
          \bottomrule[1.5pt]
    \end{tabular}
    \caption{Overall performance evaluation and statistical significance comparison (\textbf{mean $\pm$ std}).}
    \label{tab:wholeset}
\end{table}

\section{Evaluation of VLM Choices}

We evaluate several state-of-the-art VLMs as potential backbones for our experiments. The candidates include GPT-4o, Gemini 2.0 Flash, Pixtral-Large, Qwen-Omni-Turbo, and LLaVa-1.5-13B. Each model is tested using a standardized pipeline with frontier images as input, without additional fine-tuning or task-specific adaptation. The performance, reported in terms of Success Rate (SR) and Success weighted by Path Length (SPL), is summarized in Table~\ref{tab:vlmchoice}.

Among the candidates, GPT-4o achieves the highest performance, outperforming others significantly in both SR and SPL. Therefore, we adopt GPT-4o as our default VLM throughout the experiments.

\begin{table}[h]
    \centering
    \begin{tabular}{lcc}
    \toprule[1.5pt]
    VLM & SR & SPL \\
    \midrule[0.75pt]
    LLaVa-1.5-13B & -- & -- \\
    Pixtral-Large & -- & -- \\
    Qwen-Omni-Turbo & 17.27 & 1.10 \\
    Gemini-2.0-Flash & 59.35 & 43.36 \\
    GPT-4o & \textbf{62.95} & \textbf{48.34} \\
    \bottomrule[1.5pt]
    \end{tabular}
    \caption{Comparison of different VLMs for frontier potential estimation. ``--'' indicates the model lacks sufficient multimodal understanding or compatibility in our setting.}
    \label{tab:vlmchoice}
\end{table}

\section{Significance Evaluation Statistics}

We also provide the detailed statistics of the significance evaluation. On the GOAT-Bench benchmark, SCOPE achieves a mean success rate of 70.14\% ($\pm$ 1.88), outperforming 3D-Mem, which achieves 65.47\% ($\pm$ 4.02). On A-EQA, SCOPE attains a correctness of 52.37\% ($\pm$ 1.50) compared to 49.47\% ($\pm$ 3.62) from 3D-Mem. 

We apply an unpaired two-tailed t-test to assess the statistical significance of these improvements. The resulting p-value on GOAT-Bench is 0.046, indicating that the improvement is statistically significant at the 5\% level. On A-EQA, the p-value is 0.1365, suggesting a positive trend in favor of SCOPE.
These results indicate that SCOPE not only achieves higher average performance but also exhibits lower variance across runs. This highlights the robustness and consistency of its improvements over 3D-Mem. Together with the calibration analysis, the findings support SCOPE as a more stable and effective agent for embodied navigation tasks.

\section{Calibration Evaluation Statistics}

To evaluate the reliability of SCOPE, we perform a calibration analysis on both GOAT-Bench and A-EQA. The statistic result is shown in Table~\ref{tab:calibration}, SCOPE achieves superior calibration performance. On GOAT-Bench, it reduces ECE from 11.6 (of the most competitive baseline 3D-Mem) to 3.8, and on A-EQA, from 11.6 to 8.1. These results indicate that SCOPE not only improves task performance but also produces more trustworthy confidence estimates, equipping the agent with more reliable self-assessment for acting under uncertainty.

\begin{table}[h]
    \centering
    \begin{tabular}{ccccc}
         \toprule[1.5pt]
         \multirow{2}{*}{Bin} & \multicolumn{2}{c}{GOAT-Bench Acc} & \multicolumn{2}{c}{A-EQA Acc} \\
           & 3D-Mem & SCOPE & 3D-Mem & SCOPE \\
          \midrule[0.75pt]
          0-10   & 0.0   & 0.0  & 0.0 & 0.0 \\ 
          10-20  & 100.0 & 0.0  & 0.0 & 0.0 \\ 
          20-30  & 50.0  & 0.0  & 0.0 & 0.0 \\ 
          30-40  & 33.3  & 25.0 & 0.0 & 0.0 \\ 
          40-50  & 50.0  & 0.0  & 0.0 & 0.0 \\ 
          50-60  & 0.0   & 50.0 & 75.0 & 0.0 \\ 
          60-70  & 0.0   & 0.0  & 0.0 & 0.0 \\ 
          70-80  & 25.0  & 71.4 & 57.8 & 25.0 \\ 
          80-90  & 74.2  & 84.1 & 80.4 & 80.0 \\ 
          90-100 & 92.6  & 87.5 & 0.0 & 100.0 \\ 
          ECE    & 11.6  & 3.8  & 11.6 & 8.1 \\
          
          \bottomrule[1.5pt]
    \end{tabular}
    \caption{Calibration evaluation performance comparison.}
    \label{tab:calibration}
\end{table}

\section{Case Study}

\begin{figure}[h]
    \centering
    \includegraphics[width=0.5\columnwidth]{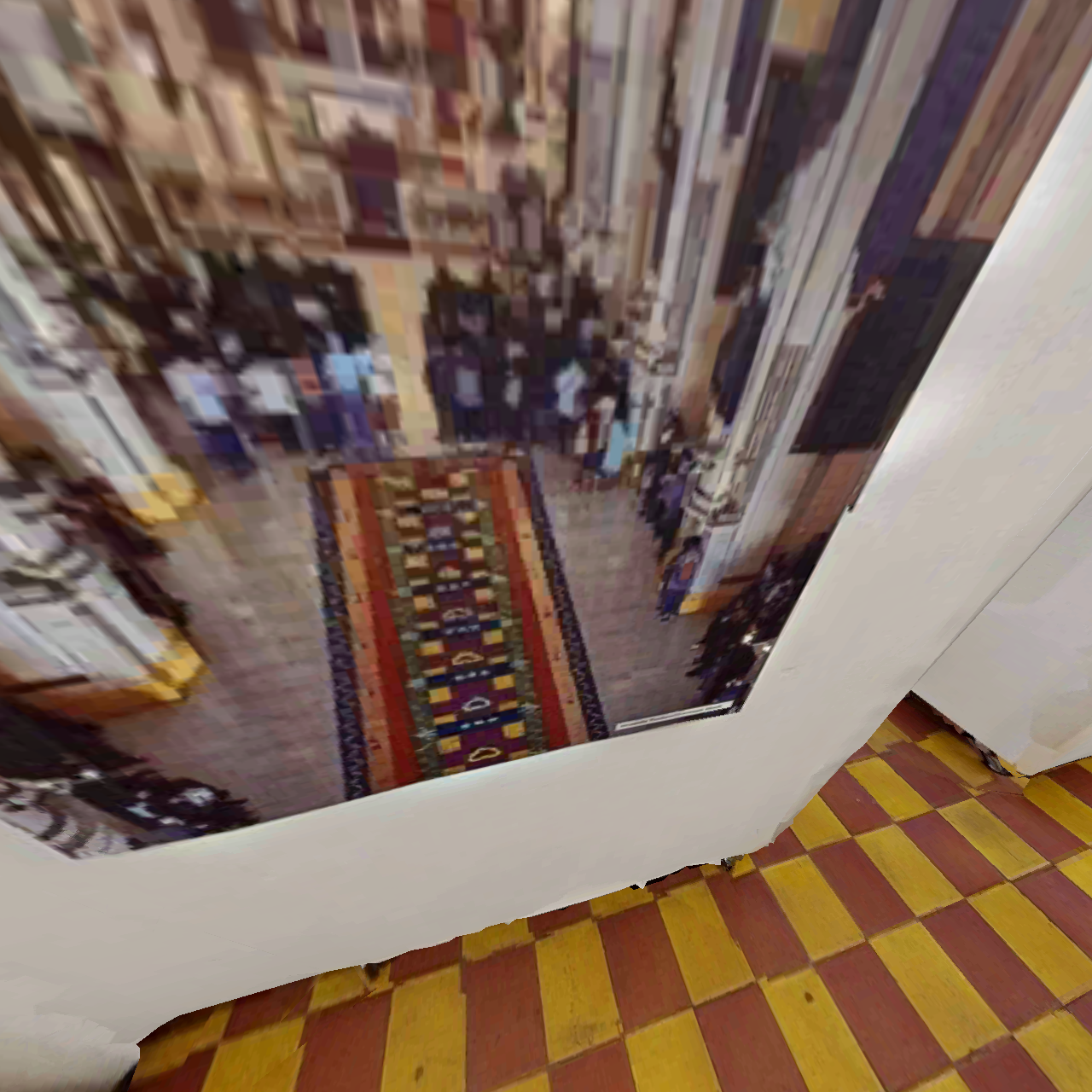}
    \caption{Goal image.}
    \label{fig:goal}
\end{figure}

We conduct a case study to examine the agent's ability to locate a specific object based on visual and contextual cues. The task is to find the exact object depicted at the center of a provided goal image, shown in Fig.~\ref{fig:goal}. The associated prompt is: ``Could you find the exact object captured at the center of the following image? You need to pay attention to the environment and find the exact object.''

\begin{figure}[h]
    \centering
    \includegraphics[width=0.95\columnwidth]{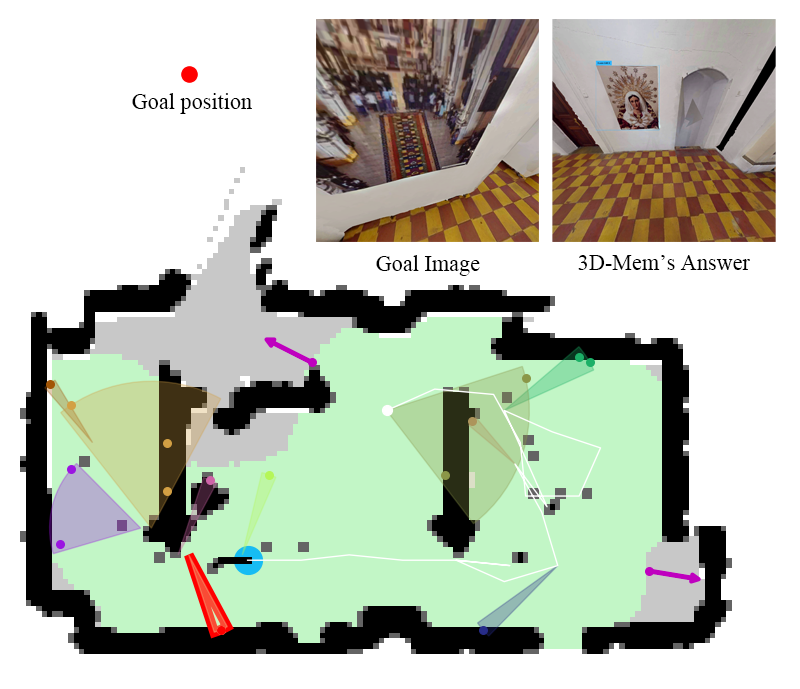}
    \caption{Navigation path of 3D-Mem. The agent remains confined to the initial room and fails to locate the actual goal object (red dot).}
    \label{fig:3dmem}
\end{figure}

We compare the performance of 3D-Mem and our proposed method, SCOPE. As illustrated in Fig.~\ref{fig:3dmem}, 3D-Mem fails in all five times attempts. The agent spends substantial time wandering within the same room (path indicated by the white line), failing to realize that the goal object (red dot) is located in a different room. Due to the lack of structured memory and scene-level reasoning, the agent engages in aimless exploration and ultimately selects a visually similar but incorrect image as the goal.

\begin{figure}[h]
    \centering
    \includegraphics[width=0.95\columnwidth]{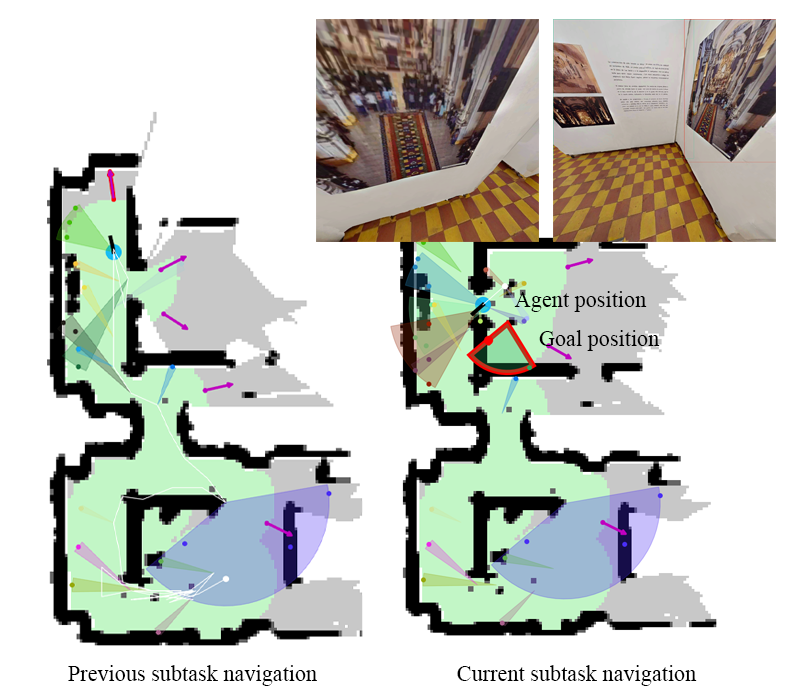}
    \caption{Navigation path of SCOPE. Leveraging prior knowledge, the agent directly returns to the room containing the goal object and successfully finds the goal picture.}
    \label{fig:scope}
\end{figure}

In contrast, as shown in Fig.~\ref{fig:scope}, SCOPE successfully identifies the goal object immediately after the task begins. During a previous subtask, the agent had already explored the correct room and stored a snapshot of the goal object. Leveraging its structured memory and potential-aware planning, the agent efficiently recalls this snapshot and navigates directly to the goal location without unnecessary exploration.



\end{document}